\documentclass[11pt,a4paper]{article}
\usepackage[hyperref]{emnlp2018}
\usepackage{times}
\usepackage{latexsym}
\usepackage{graphicx} 
\usepackage{todonotes} 
\usepackage{soul} 
\usepackage{url}
\usepackage{CJKutf8} 

\usepackage{xcolor}
\colorlet{purple}{black}
\colorlet{red}{black}
\colorlet{blue}{black}

\aclfinalcopy 

\title{Attaining the Unattainable? Reassessing Claims\\of Human Parity in Neural Machine Translation} 

\author{Antonio Toral \\
  Center for Language and Cognition\\
  University of Groningen\\
  The Netherlands\\
  {\tt \small{a.toral.ruiz@rug.nl}} \\\And
  Sheila Castilho ~~~~~ Ke Hu ~~~~~ Andy Way\\
  ADAPT Centre\\
  Dublin City University\\
  Ireland\\
  {\tt \small{firstname.secondname@adaptcentre.ie}}\\}

\date{}

\begin{document}
\maketitle
\begin{abstract}
We reassess a recent study~\cite{achieving-human-parity-on-automatic-chinese-to-english-news-translation} that claimed that machine translation (MT) has reached human parity for the translation of news from Chinese into English, using pairwise ranking and considering three variables that were not taken into account in that previous study: the language in which the source side of the test set was originally written, the translation proficiency of the evaluators, and the provision of inter-sentential context.
If we consider only original source text (i.e. not translated from another language, or  translationese)
, then we find evidence showing that human parity has not been achieved.
We compare the judgments of professional translators against those of non-experts and discover that those of the experts result in higher inter-annotator agreement and better discrimination between human and machine translations.
In addition, we analyse the human translations of the test set and identify important translation issues.
Finally, based on these findings, we provide a set of recommendations for future human evaluations of MT.
\end{abstract}

\section{Introduction}


Neural machine translation (NMT) has revolutionised the field of MT by overcoming many of the weaknesses of the previous state-of-the-art phrase-based machine translation (PBSMT)~\citep{D16-1025,toral-sanchezcartagena:2017:EACLlong}. 
In only a few years since the first working models, this approach has led to a substantial improvement in translation quality, reported in terms of automatic metrics~\cite{bojar-EtAl:2016:WMT1,bojar-EtAl:2017:WMT1,sennrich-wmt16}. This has ignited higher levels of expectation, fuelled in part by hyperbolic claims from large MT developers. First we saw in \citet{Google_NMT_16} that Google NMT was ``bridging the gap between human and machine translation [quality]”. This was amplified recently by the claim by \citet{achieving-human-parity-on-automatic-chinese-to-english-news-translation} that Microsoft had "achieved human parity” in terms of translation quality on news translation from Chinese to English, and more recently still by SDL who claimed to have ``cracked" Russian-to-English NMT with ``near perfect" translation quality.\footnote{https://www.sdl.com/about/news-media/press/2018/sdl-cracks-russian-to-english-neural-machine-translation.html} 
However, when human evaluation is used to compare NMT and SMT, the results do not always favour NMT \cite{castilho2017neural,Castil-MTSummit2017}.
 
Accompanying the claims regarding the capability of the Microsoft Chinese-to-English NMT system, \citet{achieving-human-parity-on-automatic-chinese-to-english-news-translation} released their experimental data\footnote{http://aka.ms/Translator-HumanParityData} which permits replicability of their experiments. In this paper, we provide a detailed examination of Microsoft's claim to have reached {\it human parity} for the task of translating news from Chinese (ZH) to English (EN). They provide two definitions in this regard, namely:

{\bf Definition 1}. {\em If a bilingual human judges the quality of a candidate translation produced by a
human to be equivalent to one produced by a machine, then the machine has achieved human
parity.}

{\bf Definition 2}. {\em If there is no statistically significant difference between human quality scores for a
test set of candidate translations from a machine translation system and the scores for the corresponding
human translations then the machine has achieved human parity.}

The remainder of the paper is organised as follows. First, we identify and discuss three potential issues
in Microsoft's human evaluation, concerning (i) the language in which the source text was originally written, (ii) the competence of the human evaluators with respect to translation, and (iii) the linguistic context available to these evaluators (Section \ref{s:issues}). 
We then conduct a new modified evaluation of their MT system on the same dataset taking these issues onboard {(Section \ref{s:evaluation}). In so doing, we reassess whether human parity has indeed been achieved following what we consider to be a fairer evaluation setting.
We then take a closer look at the  quality of Microsoft's dataset with the help of an English native speaker and a Chinese native speaker, and discover a number of problems in this regard (Section \ref{s:analyses}). Finally, we conclude the paper (Section~\ref{conc}) with a set of recommendations for future human evaluations, together with some remarks on the risks for the whole field of overhyping the capability of the systems we build.

\section{Potential Issues}\label{s:issues}

\subsection{Original Language of the Source Text}

The test set used by~\newcite{achieving-human-parity-on-automatic-chinese-to-english-news-translation} (\texttt{newstest2017}) was the ZH reference from the news translation shared task at WMT 2017~\cite{bojar-EtAl:2017:WMT1},\footnote{{http://www.statmt.org/wmt17/translation-task.html}} which contains 2,001 sentence pairs, of which half were originally written in ZH and the remaining half were originally written in EN. 
Figure~\ref{fig:test set} represents the WMT test set and the respective translation. 
The organisers of WMT 2017 manually translated each of these two subsets (files A1 and B1 in Figure~\ref{fig:test set}) into the other language (B2 and A2, respectively) to produce the resulting parallel test set of 2,001 sentence pairs. Thus, \newcite{achieving-human-parity-on-automatic-chinese-to-english-news-translation} machine-translated 2,001 sentences from ZH into EN, but only half of them were originally written in ZH (file D1); the other half were originally written in EN, then they were translated by a human translator into ZH (as part of WMT's organisation), and this human translation was finally machine-translated by Microsoft into EN (file D2). Microsoft also human-translated the ZH reference file into EN to use as reference translations (file C - EN REF). Therefore, 50\% of their EN reference comprises EN translations direct from the original Chinese (file C1), while 50\% are EN translations from the human-translated file from EN into ZH (file C2), i.e. backtranslation of the original EN (A1). 
While their human evaluation is conducted on three different subsets (referred to as Subset-2, Subset-3, and Subset-4 in Tables 5d to 5f of their paper), since all three are randomly sampled from the whole test set, these subsets still contain 
around 50\% of sentences originally written in ZH and around 50\% originally written in EN.

We hypothesize that the sentences originally written in EN are easier to translate than those originally written in ZH, due to the simplification principle of translationese, namely that translated sentences tend to be simpler than their original counterparts \citep{laviosa1998universals}. Two additional universal principles of translation, explicitation and normalisation, would also indicate that a ZH text originally written in EN would be easier to translate. Therefore, we explore whether the inclusion of source ZH sentences originally written in EN distorts the results, and unfairly favours MT.


\begin{figure}[t]
\includegraphics [width=.5\textwidth]{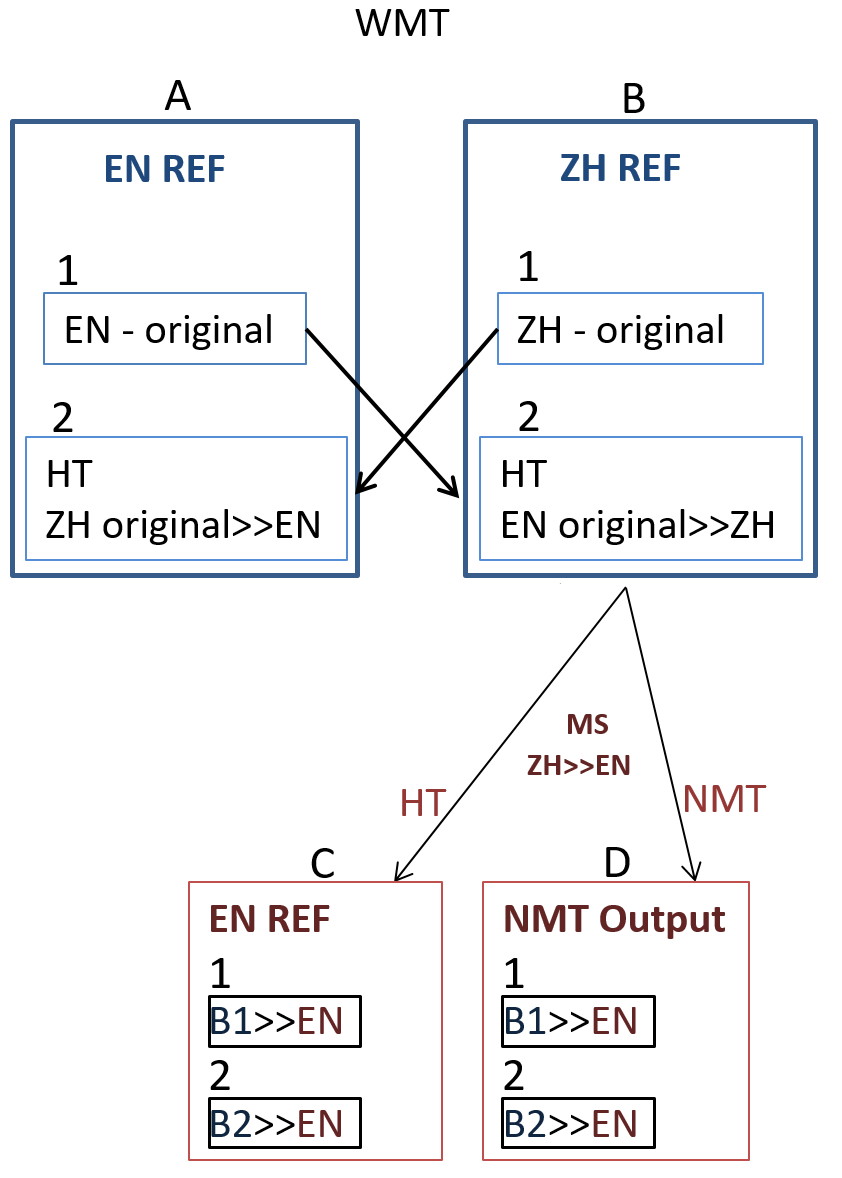}
\caption{WMT test set and Microsoft Translation ZH-to-EN reference and MT output}
\label{fig:test set} 
\end{figure}

\subsection{Human Evaluators}

The human evaluation described in~\newcite{achieving-human-parity-on-automatic-chinese-to-english-news-translation} was conducted by “bilingual crowd workers”. While the authors implemented a set of quality controls to “guarantee high quality results”, no further details are provided on the selection of evaluators and their linguistic expertise. In addition, no inter-annotator agreement (IAA) figures were provided.
We acknowledge, however, that agreement cannot be measured using the conventional Kappa coefficient, since their human evaluation uses a continuous scale (range $[0-100]$).

It has been argued that non-expert translators 
 lack knowledge of translation and so might not notice subtle differences that make one translation better than another. This was observed in the human evaluation of the TraMOOC project\footnote{http://tramooc.eu/} in which authors compared the evaluation of MT output of professional translators against \st{the} crowd workers \citep{TC39}. Results showed that for all language pairs (involving 11 languages), the crowd workers tend to be more accepting of the MT output by giving higher fluency and adequacy scores and performing very little post-editing. 

With that in mind, we attempt to replicate the results achieved in \citet{achieving-human-parity-on-automatic-chinese-to-english-news-translation} by redoing the manual evaluation with participants with different levels of translation proficiency, namely professional translators (henceforth referred to as experts) and bilingual speakers with no formal translation qualifications (henceforth referred to as non-experts).


\subsection{Context}

\newcite{achieving-human-parity-on-automatic-chinese-to-english-news-translation} evaluated the sentences in the testset in randomised order, meaning that sentences were evaluated in isolation.
However, documents such as the news stories that make up the test set contain relations that go beyond the sentence level. To translate them correctly one needs to take this inter-sentential context into account~\cite{W12-2503,Wang:2017:EMNLP}. The MT system by~\newcite{achieving-human-parity-on-automatic-chinese-to-english-news-translation} translates sentences in isolation while humans naturally consider the wider context when conducting translation.

Our hypothesis is that referential relations that go beyond the sentence-level were ignored in the evaluation as its setup considered sentences in isolation (randomised). This probably resulted in the evaluation missing some errors by the MT system that might have been caused by its lack of inter-sentential contextual knowledge.
In contrast, our revised human evaluation takes inter-sentential context into account. Sentences are not randomised but evaluated in the order they appear in the documents that make up the test set. In addition, when a sentence is evaluated, the evaluator can see both the previous and the next sentence, akin to how a professional translator works in practice.
In the same spirit, concurrent work by \citet{laeubli2018parity} 
contrasts the evaluation of single sentences and entire documents in the dataset by~\citet{achieving-human-parity-on-automatic-chinese-to-english-news-translation}, and shows a stronger preference for human translation over MT
when evaluating documents as compared to isolated sentences.

\section{Evaluation}\label{s:evaluation}

\subsection{Experimental Setup}

We conduct a human 
evaluation in which at the same time evaluators are shown a source ZH sentence and three EN translations thereof: (i) the human translation produced by Microsoft (file C in Figure \ref{fig:test set}: henceforth referred to as HT), (ii) the output of Microsoft's MT system (file D: henceforth MS), and the output of a production system, Google Translate (henceforth GG).\footnote{\textcolor{red}{We note that in the study by \citet{achieving-human-parity-on-automatic-chinese-to-english-news-translation}, 9 different translations were compared: 3 reference translations, and the output from six MT systems, 4 of which were Microsoft systems (including one online), plus Google Translate and the Sogou system \cite{sogou}, the best-performing system at WMT-2017. This, together with the fact that we use different methods, may affect the comparability of the results obtained to some degree.}}
We take these three translations from the data provided by~\citet{achieving-human-parity-on-automatic-chinese-to-english-news-translation}.

Instead of giving evaluators randomly selected sentences, they see them in order. We randomised the documents in the test set (169) and prepared one evaluation task per document, for the first 49 documents (503 sentences). 
Of these 49 documents, 41 were originally written in ZH (amounting to 299 sentences, with each document containing 7.3 sentences on average) and the remaining 8 were originally written in EN (204 sentences, average of 25.5 sentences per document).
Evaluators were asked to annotate all the sentences of each document in one go, so that they can take intersentential context into account.


\begin{figure*}[htbp]
\includegraphics[width=\textwidth]{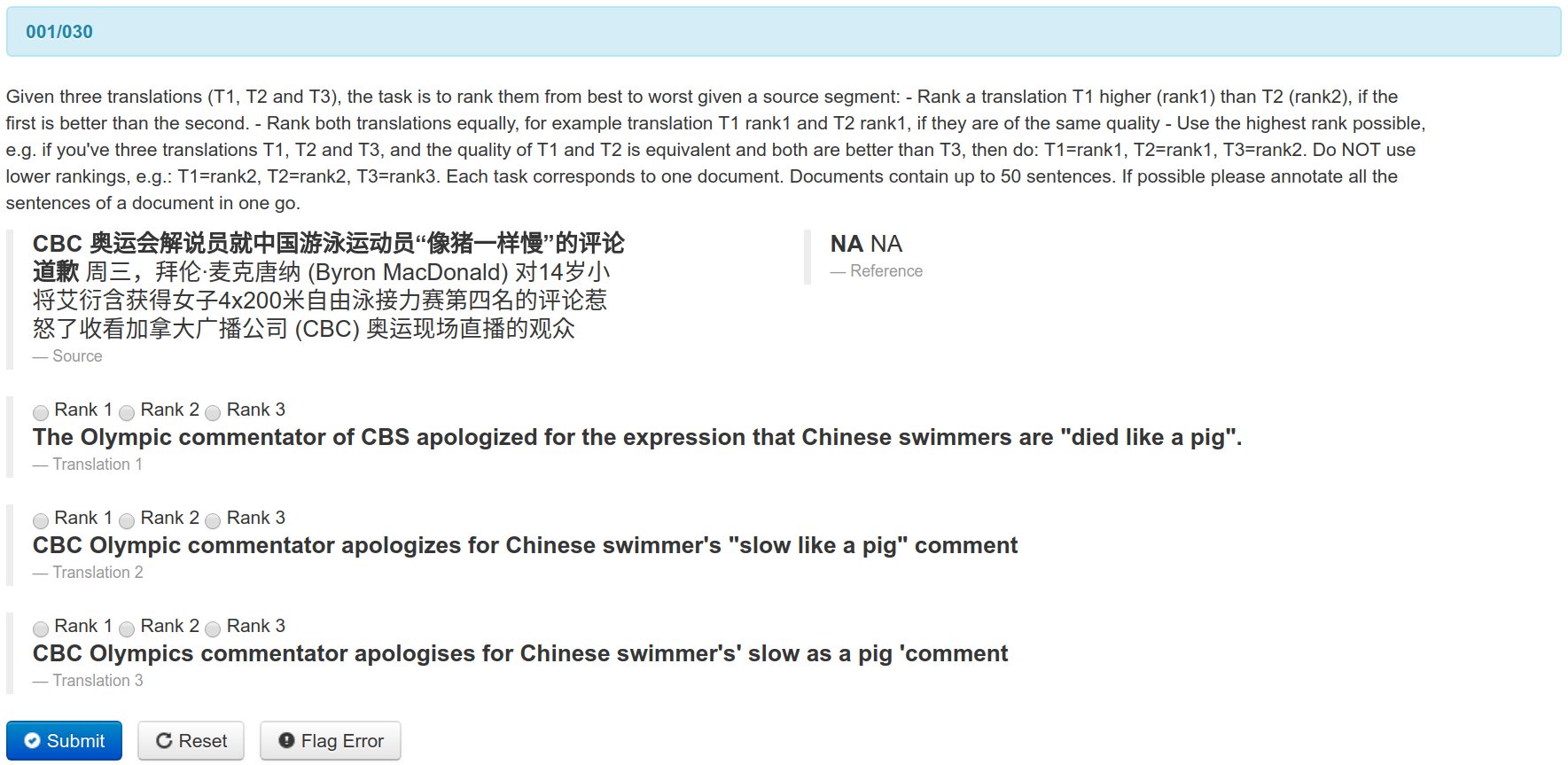}
\caption{Snapshot from the human evaluation showing the first sentence from the first document, which contains 30 sentences.}
\label{f:appraise_snapshot}
\end{figure*}

Rather than direct assessment (DA)~\cite{N15-1124}, as in~\citet{achieving-human-parity-on-automatic-chinese-to-english-news-translation}, we conduct a relative ranking evaluation. While DA has some advantages over ranking and has replaced the latter at the WMT shared task since 2017~\cite{bojar-EtAl:2017:WMT1}, ranking is more appropriate for our evaluation due to the fact that we evaluate sentences in consecutive order (rather than randomly). This can be accommodated in ranking as we can show all three translations for each source sentence together with the previous and next source sentences at the same time. In contrast, in DA only one translation is shown at a time, which is of course evaluated in isolation. An important advantage of DA is that the number of annotations required grows linearly (rather than exponentially with ranking) with the number of translations to be evaluated; this is relevant for WMT's shared task as there may be many MT systems to be evaluated, but not for our research as we have only three translations (HT, MS and GG). In any case, both approaches have been found to lead to very similar outcomes as their results correlate very strongly ($R\geq 0.92$ in~\citet{bojar-EtAl:2016:WMT1}).

Our human evaluation is performed with the Appraise tool~\citep{mtm12_appraise}.\footnote{https://github.com/cfedermann/Appraise}
Figure~\ref{f:appraise_snapshot} shows a snapshot of the evaluation.
Subsequently, we derive an overall score for each translation (HT, MS and GG) based on the rankings.
To this end we use the TrueSkill method adapted to MT evaluation~\citep{sakaguchi-post-vandurme:2014:W14-33} 
following its usage at WMT15,\footnote{https://github.com/mjpost/wmt15}
i.e. we run 1,000 iterations of the rankings recorded with Appraise followed by clustering (significance level $\alpha=0.05$).

Five evaluators took part in our evaluation: two professional Chinese-to-English translators 
and three non-experts. 
Of the two professional translators, one is a native English speaker with a fluent level of Chinese, and the other  is a Chinese native speaker with a fluent level of English.
The three non-expert bilingual participants are Chinese native speakers with an advanced level of English.
These bilingual participants are researchers in NLP, and so their profile is similar to some of the human evaluators of WMT, namely MT researchers.\footnote{\textcolor{red}{It is an open question as to whether using bilingual NLP researchers may affect the results obtained. While we follow the practice of WMT here -- which differs from the approach taken by \citet{achieving-human-parity-on-automatic-chinese-to-english-news-translation}, who used bilingual crowd workers -- we intend in future work to investigate this further.}}

All evaluators completed all  49 documents, except the third non-expert, who completed the first 18. Similarly, all evaluators ranked all the sentences in the documents they evaluated, except the second professional translator, who skipped 3 sentences.
In total we  collected 6,675 pairwise judgements.

\subsection{Results}

\subsubsection{Original Language}

To find out whether the language in which the source sentence was originally written has any effect on the evaluation,
we show the resulting Trueskill scores for each translation taking into account all the sentences in our test set versus considering the sentences in two groups according to the original language (ZH and EN).
The results are shown in Table \ref{t:orig_lang}.

\begin{table}[htbp]
\begin{center}
\begin{tabular}{|l|l|l|l|}
\hline
\bf Rank & \multicolumn{3}{c|}{\bf Original language}\\
\hline
		 & \multicolumn{1}{c|}{\bf Both} 		& \multicolumn{1}{c|}{\bf ZH}	& \multicolumn{1}{c|}{\bf EN}\\
		 & $n=6675$		& $n=3873$	& $n=2802$\\
\hline
1		 & HT 1.587*	& HT 1.939*	& MS 1.059\\
2		 & MS 1.231*	& MS 1.199*	& HT 0.772*\\
3		 & GG -2.819	& GG -3.144 & GG -1.832\\
\hline
\end{tabular}
\end{center}
\caption{\label{t:orig_lang} Ranks of the translations given the original language of the source side of the test set shown with their Trueskill score (the higher the better). An asterisk next to a translation indicates that this translation is significantly better than the one in the next rank.
}
\end{table}

Regardless of the original language, GG is the lowest-ranked translation, thus providing an indication that the quality obtainable from the MS system is a notable improvement over state-of-the-art NMT systems used in production.
We observe that HT outperforms significantly MS when the original language is ZH, but the difference between the two is not significant when the original language is EN.
Hence, we confirm our hypothesis that the use of translationese as the source language distorts the results in favour of MS.

Next, we check whether this effect of translationese is also present in the evaluation by~\citet{achieving-human-parity-on-automatic-chinese-to-english-news-translation}.
To this end, we concatenate all their judgments and model them with mixed-effects beta regression.
Our dependent variable is the score, scaled down from the original range $[0,100]$ to $[0,1]$, which we aim to predict with one continuous predictor -- sentence length -- and two factorial independent variables: translation (levels HT, MS and GG) and original language (levels ZH and EN).
The identifiers of the sentence and the annotator are included as random effects.
We plot the interaction between the translation and the original language of the resulting model (adjusted $R^2=0.32$) in Figure~\ref{f:systemid_origlang}.
HT outperforms MS by around 0.05 absolute points 
($p=0.06$) for sentences whose original language is ZH. However this gap disappears for source sentences originally written in EN, where we see that the score for MS is actually slightly higher than that of HT, though the difference is not significant \textcolor{blue}{($p=0.3$)}.
We observe a clear effect of translationese (EN): compared to ZH, the scores of both MT systems increase substantially (GG over 10\% absolute and MS over 6\% absolute), while the HT score increases only very slightly.
\textcolor{blue}{
We then build the same regression model for the subset of judgments whose source text was originally written in ZH. The difference between HT and MS is significant ($p<0.05$) in favour of the first in the resulting model (adjusted $R^2=0.36$).}

\begin{figure}[htbp]
\includegraphics[width=0.49\textwidth]{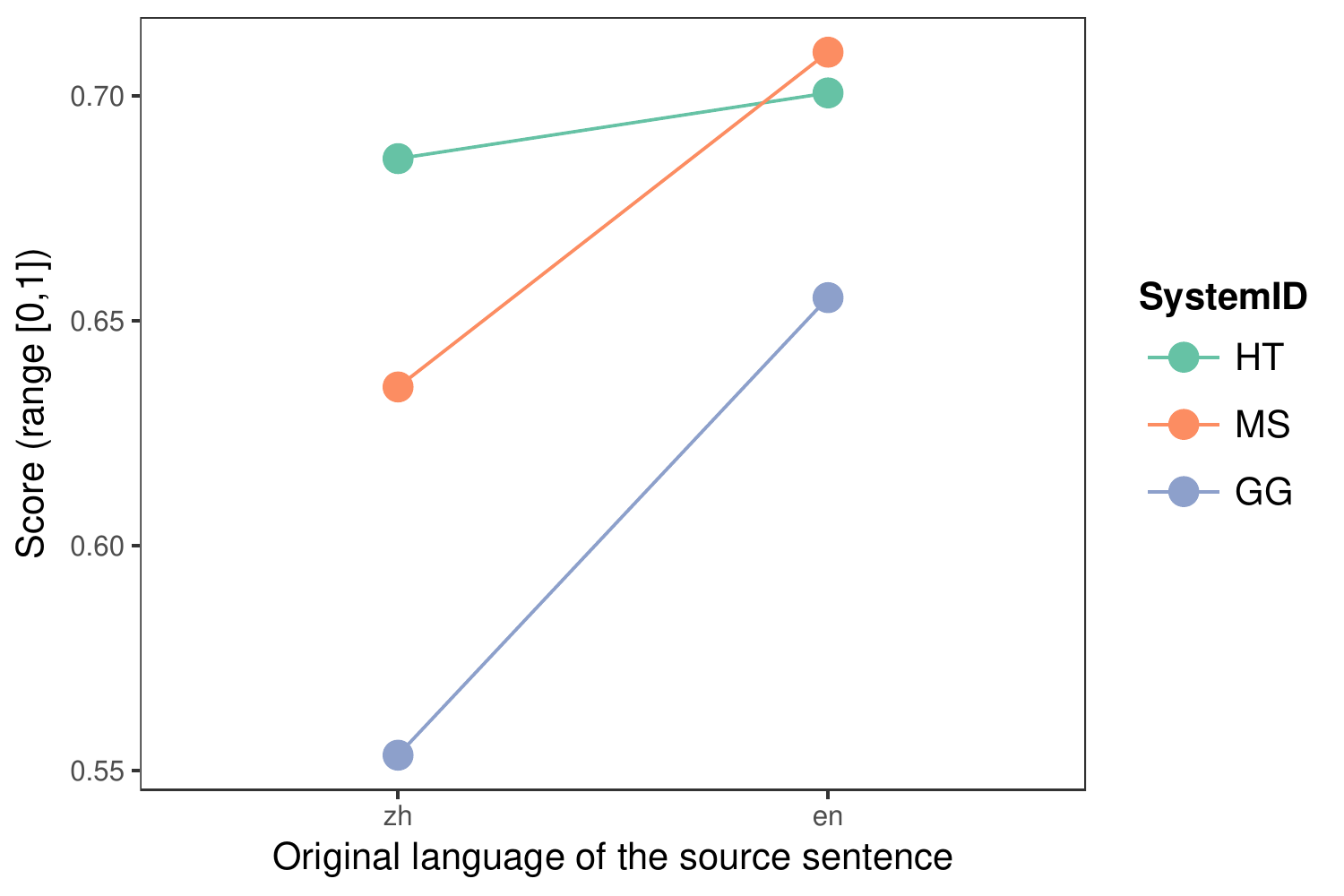}
\caption{Interaction between the MT system (levels HT, MS and GG) and the original language of the source sentence (levels ZH and EN).}
\label{f:systemid_origlang}
\end{figure}

Our hypothesis was theoretically supported by the simplification principle of translationese.
Applied to the test data, this would mean that the portion originally written in ZH is more complex than the part originally written in EN.
To check whether this is the case, we compare the two subsets of the test set using a measure of text complexity, type-token ratio (TTR). 
While both subsets contain a similar number of sentences (1,001 and 1,000), the ZH subset contains more tokens (26,468) than its EN counterpart (22,279). We thus take a subset of the ZH (840 sentences) containing a similar amount of words to the EN data (22,271 words).
We then calculate the TTR for these two subsets using bootstrap resampling.
The TTR for ZH ($M=0.1927$, $SD=0.0026$, 95\% confidence interval $[0.1925,0.1928]$) is 13\% higher than that for EN ($M=0.1710$, $SD=0.0025$, 95\% confidence interval $[0.1708,0.1711]$).

Given the findings of this experiment, in the remainder of the paper we use only the subset of the test set whose original language is ZH.

\subsubsection{Evaluators}

To find out whether the translation expertise of the evaluator has any effect on the evaluation,
we show the resulting Trueskill scores for each translation resulting from the evaluations by non-expert versus expert translators.
The results are shown in Table \ref{t:evaluators}.
The gap between HT and MS is considerably wider for experts (2.2 vs 1.2) than for non-experts (1.3 vs 0.9).
We link this to our expectation, based on the previous finding by \citet{TC39}, that non-experts are more lenient regarding MT errors.
In other words, non-experts disregard translation subtleties in their evaluation, which leads to the gap between different translations -- in this case HT and MS -- being smaller.
In Section \ref{s:analyses} we  explore this further by means of a qualitative analysis.

\begin{table}[htbp]
\begin{center}
\begin{tabular}{|l|l|l|l|}
\hline
\bf Rank & \multicolumn{3}{c|}{\bf Translators}\\
\hline
		 & \multicolumn{1}{c|}{\bf All} 		& \multicolumn{1}{c|}{\bf Experts} 	& \multicolumn{1}{c|}{\bf Non-experts}\\
         & $n=3873$				& $n=1785$				& $n=2088$\\
\hline
1		 & HT 1.939*	& HT 2.247*		& HT 1.324\\
2		 & MS 1.199*	& MS 1.197*		& MS 0.94*\\
3		 & GG -3.144 	& GG -3.461		& GG -2.268\\
\hline
\end{tabular}
\end{center}
\caption{\label{t:evaluators}
Ranks and Trueskill scores (the higher the better) of the three translations for evaluations carried out by expert versus non-expert translators. An asterisk next to a translation indicates that this translation is significantly better than the one in the next rank. }
\end{table}

Trueskill provides not only an overall score for each translation but also its confidence interval.
We expect these to be
wider for the annotations by non-experts than those annotations given by experts, which
would indicate that there is more uncertainty in the rankings by non-experts.
Figure~\ref{f:cis} shows the scores for each translation by experts and non-experts, i.e. the same values that were shown in Table~\ref{t:evaluators}, now enriched with their 95\% confidence intervals.

The sum of the confidence scores for the three translations is just 0.33\% higher for non-experts (3.076) than for experts (3.066).
However, it is worth mentioning that, compared to the width of the intervals for experts, those for non-experts are considerably wider for HT (16\% relative difference) while they are similar or smaller for MT (1\% and -11\% relative differences for GG and MS, respectively).


\begin{figure}[htbp]
\includegraphics[width=0.5\textwidth]{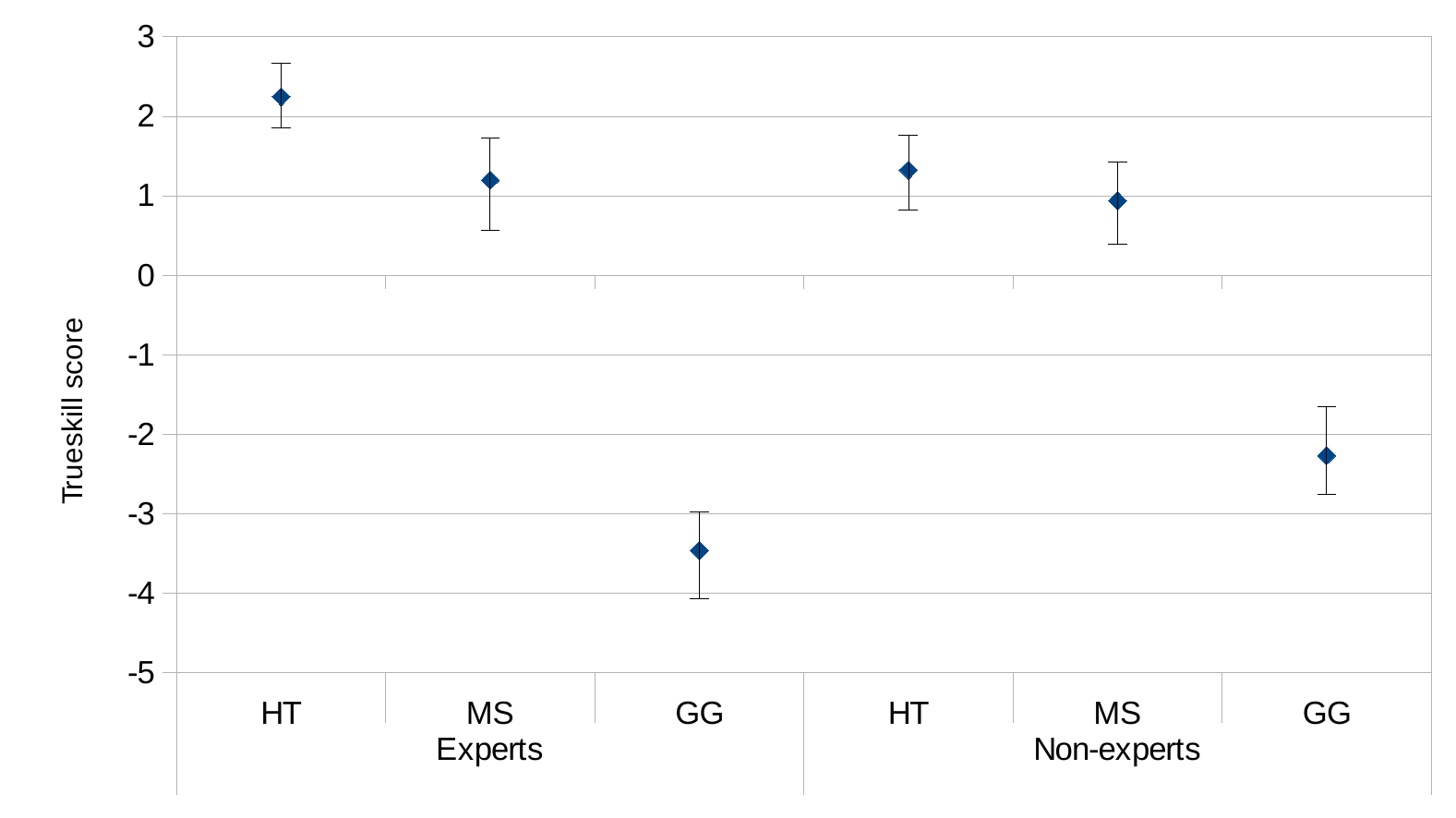}
\caption{Trueskill scores of the three translations by experts and non-experts together with their confidence intervals.}
\label{f:cis}
\end{figure}

We now look at inter-annotator agreement (IAA) between experts and non-experts.
We compute the Kappa ($\kappa$) coefficient~\cite{cohen_doi:10.1177/001316446002000104}, as done at WMT 2016~\cite[Section~3.3]{bojar-EtAl:2016:WMT1}:\footnote{\url{https://github.com/cfedermann/wmt16/blob/master/scripts/compute_agreement_scores.py}}
\[
k = \frac{P(A) - P(E)}{1- P(E)}
\]
\noindent
where $P(A)$ represents the proportion of times that the annotators agree, and $P(E)$ the proportion of times that the annotators are expected to agree by chance.

As expected, the IAA between professional translators ($\kappa=0.254$) is notably higher, 95\% relative, than that between non-experts ($\kappa=0.130$).\footnote{Due to the fact that one non-expert evaluated only 18 out of the 49 documents, the IAA calculations consider only the first 18 documents. If we consider all  49 documents, the trend remains the same; the IAA for the two experts is higher than that for the two non-experts who evaluated all the documents: 0.265 vs 0.196.}
As we have three non-experts, we can calculate the IAA not only among the three of them but also between all three pairs of non-expert annotators; all of the resulting coefficients (0.057, 0.135 and 0.195) are lower than that between experts (0.254).

To the best of our knowledge, this is the first time that IAA of professional translators and non-experts has been compared for the human evaluation of MT.
In related work, 
\citet{Callison-Burch:2009:FCC:1699510.1699548} compared the agreement level of two types of non-expert translators: MT developers (referred to in that paper as `experts') and crowd workers.
He showed that crowd workers can reach the agreement level of MT researchers using multiple workers and weighting their judments.
That said, both types of non-experts conducted human evaluations for WMT13~\cite{bojar-EtAl:2013:WMT} and the IAA rates of the crowd were well below those of the researchers.


\section{Analyses}\label{s:analyses}

As mentioned previously, we have examined the quality of the test sets, both originally written in ZH and originally written in EN and their respective translations. An English native speaker analysed both the original EN version from the WMT set (file A1 in Figure \ref{fig:test set}) and the human translation of the set originally written in ZH performed by Microsoft (file C2). 
A Chinese native speaker, who is fluent in English and has experience with translation from EN into ZH, analysed the original ZH versions (file B1) as well as the human translation of the set originally written in EN performed by the WMT organisers (file B2). 

\subsection{Original English}
Regarding the English original (file A1 in Figure \ref{fig:test set}), the analysis showed that apart from a few 
grammar errors, the test set appeared to be fluent and grammatical. Examples of grammatical errors in the original EN files are: \\

\noindent \textbf{i)} ``The idiot didn't realize they were still on the air''

\noindent \textbf{ii)} ``Soon after, Scott Russel who was hosting CBC's broadcast apologized on-air for MacDonald's comment, saying: `We apologize the comment on a swim performance made it to air.' '' \\

In example i) ``on air" should be used instead of ``on the air'', while in the example ii) a missing ``that'' should be used after ``apologize''. Nonetheless, these errors did not affect the ZH translation (file B2) or the following backtranslation (C2) into EN. Our hypothesis is that because the test set is news content, it also contains tweets (such as example i)) and quotes \st{that} from speech interviews (such as example ii)), which are more likely to contain grammatical errors.

\subsection{Chinese Translation}
Regarding the human translation into ZH performed by WMT (file B2 in Figure \ref{fig:test set}), most of the sentences contained grammatical errors and/or mistranslations of proper nouns. Furthermore, although some translations were grammatically correct and accurate, they were not fluent. When the ZH-translated sentences were compared against the source (A1), the translations were mostly accurate. However, when analyzed on their own without the source, they sound disfluent:  \\

\noindent \textbf{iii)}

\noindent EN original (A1): A front-row seat to the stunning architecture of the Los Angeles Central Library

\noindent ZH (B2):\begin{CJK}{UTF8}{gbsn}洛杉矶中央图书馆的惊艳结构先睹为快\end{CJK}
\\

\noindent \textbf{iv)}

\noindent EN original (A1): An open, industrial loft in DTLA gets a cozy makeover

\noindent ZH (B2): \begin{CJK}{UTF8}{gbsn}DTLA的开放式工厂阁楼进行了一次舒适的改造。\end{CJK}
\\

In example iii), although the ZH translation has fully transferred the meaning of the source text, it contains word-order errors which makes the translation disfluent since the verb phrase
\begin{CJK}{UTF8}{gbsn}``先睹为快''\end{CJK} (take a look) is placed after the object (library). One possible translation for that is \begin{CJK}{UTF8}{gbsn}``抢先目睹洛杉矶中央图书馆的惊艳结构''\end{CJK} because the ZH language syntax requires the verb to be placed before the object.

In example iv), the ZH translation contains a grammatical error in the word \begin{CJK}{UTF8}{gbsn}``进行''\end{CJK}, which would imply that the loft is carrying out a makeover. 
In addition, the adjective \begin{CJK}{UTF8}{gbsn}``舒适的''\end{CJK} (cosy) cannot be used to describe \begin{CJK}{UTF8}{gbsn}``改造''\end{CJK} (makeover). One possible translation for the English sentence is 
\begin{CJK}{UTF8}{gbsn} ``DTLA的开放式工业阁楼被改造的很舒适''. \end{CJK}

Given this analysis, we speculate that the translation of the EN original files into ZH might not have been performed by an experienced translator, but rather exemplify either human translation performed by an inexperienced translator, or poorly post-edited MT.

\subsection{English Translation} 
Regarding the EN reference files translated by Microsoft (file C2 in Figure \ref{fig:test set}), many of the sentences contained grammatical errors (such as  word order, verb tense and missing prepositions) as well as mistranslations. \\

\noindent \textbf{v)}
 
\noindent EN original (A1): A front-row seat to the stunning architecture of the Los Angeles Central Library

\noindent ZH (B2):\begin{CJK}{UTF8}{gbsn}洛杉矶中央图书馆的惊艳结构先睹为快\end{CJK}

\noindent EN (C2): Take a look of the astounding architecture of the Los Angeles Central Library.\\

\noindent GG: The stunning structure of the Los Angeles Central Library

\noindent MS: A sneak peek at the stunning architecture of the Los Angeles Central Library\\

\noindent \textbf{vi)}

\noindent EN original (A1): An open, industrial loft in DTLA gets a cozy makeover

\noindent ZH (B2):\begin{CJK}{UTF8}{gbsn}DTLA的开放式工厂阁楼进行了一次舒适的改造。\end{CJK}

\noindent EN (C2): A comfortable makeover was provided to the open factory building design of DTLA. \\

\noindent GG: DTLA's Open factory loft has a comfortable makeover.

\noindent MS: DTLA's open-plan factory loft has undergone a comfortable makeover.\\

In example v), the EN translation of the ZH source\footnote{It is important to note that the translators did not have access to the original EN (A1) and so the ZH file (B2) was used as the source.} analyzed previously is translated with the wrong preposition, i.e. `look of' instead of `look at'. None of the professional translators considered the reference worse than the MS output; while one translator and one non-expert considered it `as good' as the MS output, the other considered it better than MS but worse than GG. Regarding the non-expert assessment, two of them considered the HT to be as good as MS and better than GG, and one considered the HT to be worse than MS but better than GG.\\

In example vi), the EN translation (C2) of the ZH source (B2) does not have all the information expressed in ZH as the word `loft' \begin{CJK}{UTF8}{gbsn}(阁楼)\end{CJK}
is not translated. Moreover, the EN translation refers to an architectural design makeover of the building rather than an interior makeover of an attic.
Both professional translators considered the EN reference to be worse than the MS output. As far as the non-experts are concerned, two of them considered the HT to be worse than MS, while one considered it to be `as good'. This provides qualitative evidence that non-experts may be more tolerant of translation errors than professional translators.

Another example of such behaviour is the following:\\

\noindent \textbf{vii)}

\noindent EN original (A1): Learn more about the history of downtown's Central Library as the Society of Architectural Historians/Southern California Chapter hosts a salon with Arnold Schwartzman and Stephen Gee, authors of the new book "Los Angeles Central Library: A History of its Art and Architecture 

\noindent ZH (B2): \begin{CJK}{UTF8}{gbsn}美国建筑史学家学会南加利福尼亚洲分会与新书《洛杉矶中央博物馆：其艺术与建筑历史》的作者阿诺·斯瓦茨曼和史蒂芬·吉举办了一场沙龙。观众们可通过此次活动进一步了解市中心中央图书馆的历史 \end{CJK}

\noindent EN (C2):A salon will be hosted by Southern California Branch of Society of Architectural Historians and the co-authors of Los Angles Central Museum: Art and Architectural History, Arnold Schwarzman and Stephen Gee. It will deliver more knowledge of the Central Library to the participants\\

\noindent GG: The Southern California branch of the American Institute of Architectural Historians has held a salon with \begin{CJK}{UTF8}{gbsn}阿诺·斯瓦茨曼\end{CJK} and \begin{CJK}{UTF8}{gbsn}史蒂芬·吉\end{CJK}, author of the Los Angeles Central Museum: its art and architectural history. Through this event, viewers can learn more about the history of Central Library in the city centre\\
\noindent MS: The Southern California chapter of the American Society of Architectural Historians and the authors of a new book, "Los Angeles Central Museum: Its Art and Architectural History," Arnold Schwartzman and Steven Gee, hosted a salon at which viewers learned more about the history of the Central Library in the city center\\

In example vii), regarding the ZH source (B2), in addition to having the first sentence translated into past tense -- whereas the EN original (A1) shows the salon event is happening in the near future -- it also contains a typo \begin{CJK}{UTF8}{gbsn}`洲'\end{CJK} which means `continent' instead of `state' \begin{CJK}{UTF8}{gbsn}`州'\end{CJK}. Even though the typo does not affect the EN translation (C2), it shows that the quality of the ZH translation is not as high as would be expected of professional human translators. Regarding the EN translation (C2), while the first sentence is mostly fluent -- although it contains a typo in `Angles' (Angeles) and lacks the article `the' before the proper noun in the first sentence -- the second sentence lacks fluency and contains errors of omissions and mistranslations. For example, the words ``downtown'' and ``history'' \begin{CJK}{UTF8}{gbsn}(市中心\end{CJK} and \begin{CJK}{UTF8}{gbsn}历史,\end{CJK} respectively) were not transferred over to the EN reference (C2). Furthermore, the word `viewers' in the ZH translation \begin{CJK}{UTF8}{gbsn}(观众们)\end{CJK} was mistranslated as `participants'. Nonetheless, the EN translation (C2) is able to capture the correct tense of the sentence since the second sentence in the ZH translation \st{is} (B2) is ambiguous regarding verbal tense. The MS translation does a better job in keeping the fluency throughout the sentence even though it mistranslates the tense of the source in the past tense. Both professional translators assessed the HT as worse than MS, whereas two of the non-experts considered it to be as good as MS and better than GG. The third non-expert considered the HT to be worse than both MT systems. This example shows that the level of expertise of the evaluators may have an effect on the evaluation given that non-experts are wrongly more tolerant of MT errors.

Similarly to the ZH translation (B2) of the English original, we speculate that the EN translation (C2) of the ZH files is more likely a human translation performed by an inexperienced translator, or even a poorly post-edited machine translation; even if the translation was performed by an experienced translator, such that the ZH source (B2) contained errors or was disfluent, a professional translator would surely be more meticulous and fix such errors before rubber-stamping the translations.

\section{Conclusions and Future Work}
\label{conc}
This paper has reassessed a recent study that claimed that MT has reached human parity for the translation of news from Chinese into English, considering three variables that were not taken into account in that previous study: (i) the language in which the source side of the test set was originally written, (ii) the translation proficiency of the evaluators, and (iii) the provision of inter-sentential context.

The main findings of this paper are the following:
\begin{itemize}
\item If we consider the subset of the test set whose source side was originally written in ZH, there is evidence that human parity has not been achieved, i.e. the difference between the human and the machine translations is significant. This is the case both in our human evaluation and in Microsoft's.
\item Having translationese (ZH translated from EN in our study) as input, compared to having original text, results in higher scores for MT systems in Microsoft's human evaluation.
\item Compared to judgments by non-experts, those by professional translators have a higher IAA and a wider gap between human and machine translations.
\item We have identified issues in the human translations by both WMT and Microsoft. These indicate that these translations were conducted by non-experts and that were possibly post-edited MT output.
\end{itemize}

There is little doubt that human evaluation has played a very important role in MT research and development to date. As MT systems improve -- as exemplified by the progress made by \citet{achieving-human-parity-on-automatic-chinese-to-english-news-translation} over state-of-the-art production systems -- and thus the gap between them and human translators narrows, we believe that human evaluation, in order to remain useful, needs to be more discriminative. We suggest that a set of principles should be adhered to, partly based on our findings, which we outline here as recommendations:
\begin{itemize}
\item The original language in which the source side of the test sets is written should be the same as their source language. This will avoid having spurious effects because of having translationese as MT input.
\item Human evaluations should be conducted by professional translators. This allows fine-grained nuances of translations to be taken into account in the evaluation and should result in higher inter-annotator agreement.
\item Human evaluations should proceed taking the whole document into account rather than evaluating sentences in isolation. This allows for intersentential phenomena to be considered as part of the evaluation.
\item Test sets should be translated by experienced professional translators from scratch.
\end{itemize}

We are confident that adhering to these principles is sensible under any translation conditions. Of course, if the test set is faulty, then in claiming that one's MT system outperforms one's competitors, there is a risk that what one is actually demonstrating is the contrary, as if automatic evaluation metrics demonstrate a higher score, what that could be denoting is that one's output is actually closer to the faulty test set than producing better output in terms of improved translation quality {\em per se}. \textcolor{red}{Of course, this has consequences not just for the study in this paper, but for all shared tasks: past, present, and future.}\footnote{\textcolor{red}{Ideally, it would be great if multiple references were also available, but the point remains that if these are poor quality human translations, then this is likely to skew results still further.}}  

Should material be made available by Google, SDL or any other MT developers who claim `human parity' or the like, we would be very happy to apply these principles in subsequent rigorous evaluations of actual demonstrable improvements in translation quality. One thing is certain; as \citet{Way19} observes, ``those of us who have seen many paradigms come and go know that overgilding the lily does none of us any good, especially those of us who have been trying to build bridges between MT developers and the translation community for many years." We trust that our findings in this paper demonstrate that while MT quality does seem to be improving quite dramatically, human translators will continue to find gainful employment for many years to come, despite somewhat grandiose claims to the contrary.

On a final note, we acknowledge that our conclusions and recommendations are somewhat limited in that they are derived from experiments on just one language direction and five evaluators. Therefore we plan as future work to conduct similar experiments on additional language pairs with a higher number of evaluators. In the spirit of \citet{achieving-human-parity-on-automatic-chinese-to-english-news-translation}, without which this paper would not have been possible, we too make publicly available our evaluation materials, the anonymised human judgments and the statistical analyses thereof.\footnote{\url{https://github.com/antot/human_parity_mt}}

\section*{Acknowledgments}

We would like to thank the five expert and non-expert translators that took part in this study. \textcolor{red}{We are also grateful for valuable comments on this paper from Hany Hassan, the lead author on the Microsoft paper.} This research was partially supported by the iADAATPA project funded by CEF research action (2016-EU-IA-0132) under grant agreement No.1331703. The ADAPT Centre for Digital Content Technology at Dublin City University is funded under the Science Foundation Ireland Research Centres Programme (Grant 13/RC/ 2106) and is co-funded under the European Regional Development Fund.

\bibliography{wmt_hp_2018}
\bibliographystyle{acl_natbib_nourl}

\appendix

\section*{Appendix: Evaluator Instructions}
\label{s:eval_instructions}
\small
Given three translations (T1, T2 and T3), the task is to rank them from best to worst given a source segment:
\begin{itemize}
\item Rank a translation T1 higher (rank1) than T2 (rank2), if the first is better than the second.
\item Rank both translations equally, for example translation T1 rank1 and T2 rank1, if they are of the same quality.
\item Use the highest rank possible, e.g. if you've three translations T1, T2 and T3, and the quality of T1 and T2 is equivalent and both are better than T3, then do: T1=rank1, T2=rank1, T3=rank2. Do NOT use lower rankings, e.g.: T1=rank2, T2=rank2, T3=rank3.
\end{itemize}
Each task corresponds to one document. Documents contain up to 50 sentences. If possible please annotate all the sentences of a document in one go.

\end{document}